\documentclass[sigconf]{acmart}

\usepackage{array}
\usepackage{multirow}
\usepackage{amsmath}
\DeclareMathOperator*{\argmaxA}{arg\,max}
\usepackage{hyperref}

\usepackage{array}
\usepackage{multirow}
\usepackage{float}

\usepackage{array}

\usepackage{orcidlink}
\usepackage{color}

\setcopyright{acmlicensed}
\copyrightyear{2018}
\acmYear{2018}
\acmDOI{XXXXXXX.XXXXXXX}

\acmConference[Conference acronym 'XX]{Make sure to enter the correct
  conference title from your rights confirmation emai}{June 03--05,
  2018}{Woodstock, NY}
\acmISBN{978-1-4503-XXXX-X/18/06}

\copyrightyear{2024}
\acmYear{2024}
\setcopyright{acmlicensed}\acmConference[CIKM '24]{Proceedings of the 33rd ACM International Conference on Information and Knowledge Management}{October 21--25, 2024}{Boise, ID, USA}
\acmBooktitle{Proceedings of the 33rd ACM International Conference on Information and Knowledge Management (CIKM '24), October 21--25, 2024, Boise, ID, USA}
\acmDOI{10.1145/3627673.3679984}
\acmISBN{979-8-4007-0436-9/24/10}

\begin{document}

\title{Quantum Inverse Contextual Vision Transformers (Q-ICVT): A New Frontier in 3D Object Detection for AVs}


\author{Sanjay Bhargav Dharavath\,\orcidlink{0009-0009-0994-7050}} 
\affiliation{%
  \institution{Indian Institute of Technology}
  \city{Kharagpur}
  \country{India}
  }
\email{sanjaytinku810@gmail.com}
\author{Tanmoy Dam\, \orcidlink{0000-0003-3022-0971}}
\affiliation{%
  \institution{Emory University, }
  \country{ USA}}
\email{tanmoydam@yahoo.com}

\author{Supriyo Chakraborty\,\orcidlink{0000-0003-3697-0044}}
\affiliation{%
  \institution{Indian Institute of Technology}
  \city{Kharagpur}
  \country{India}}
\email{supriyochakraborty@iitkgp.ac.in}

\author{Prithwiraj Roy\, \orcidlink{0000-0002-0592-5593}}
\affiliation{%
  \institution{Global Action Alliance}
  \city{Washington}
  \country{USA}}
\email{przhr@mst.edu}

\author{Aniruddha Maiti\, \orcidlink{0000-0002-1142-6344}}
\affiliation{%
  \institution{ADP}
  \city{New Jersey}
  \country{USA}}
\email{aniruddha.maiti87@gmail.com}

\renewcommand{\shortauthors}{Trovato et al.}

\begin{abstract}

The field of autonomous vehicles (AVs) predominantly leverages multi-modal integration of LiDAR and camera data to achieve better performance compared to using a single modality. However, the fusion process encounters challenges in detecting distant objects due to the disparity between the high resolution of cameras and the sparse data from LiDAR. Insufficient integration of global perspectives with local-level details results in sub-optimal fusion performance.To address this issue, we have developed an innovative two-stage fusion process called Quantum Inverse Contextual Vision Transformers (Q-ICVT). This approach leverages adiabatic computing in quantum concepts to create a novel reversible vision transformer known as the Global Adiabatic Transformer (GAT). GAT aggregates sparse LiDAR features with semantic features in dense images for cross-modal integration in a global form. Additionally, the Sparse Expert of Local Fusion (SELF) module maps the sparse LiDAR 3D proposals and encodes position information of the raw point cloud onto the dense camera feature space using a gating point fusion approach. Our experiments show that Q-ICVT achieves an mAPH of 82.54 for L2 difficulties on the Waymo dataset, improving by 1.88\% over current state-of-the-art fusion methods. We also analyze GAT and SELF in ablation studies to highlight the impact of Q-ICVT. Our code is available at \href{https://github.com/sanjay-810/Qicvt}{\textcolor{red}{Q-ICVT}}.


\end{abstract}





\maketitle

\section{Related Work}
\textbf{LiDAR based 3D OD:} LiDAR point clouds, which are typically unordered collections of data points, can be divided into three main types: voxel-based, point-based, and point-voxel fusion methods. Voxel-based techniques, as indicated in studies like \cite{deng2021voxel,chen2022mppnet,hu2022afdetv2}, convert point cloud data into voxels and then use deep sparse convolution layers to extract features. Point-based methods, referenced in \cite{chen2022mppnet,shi2020point}, involve using stacked Multi-Layer Perceptrons (MLPs) to process raw point cloud data and obtain point-level features. Unlike earlier studies, recent works \cite{fan2022embracing,vora2020pointpainting} have introduced hybrid methods that combine point and voxel-based representations for more comprehensive feature extraction. 

\begin{figure*}
    \centering
    \scalebox{0.4}{
    \includegraphics{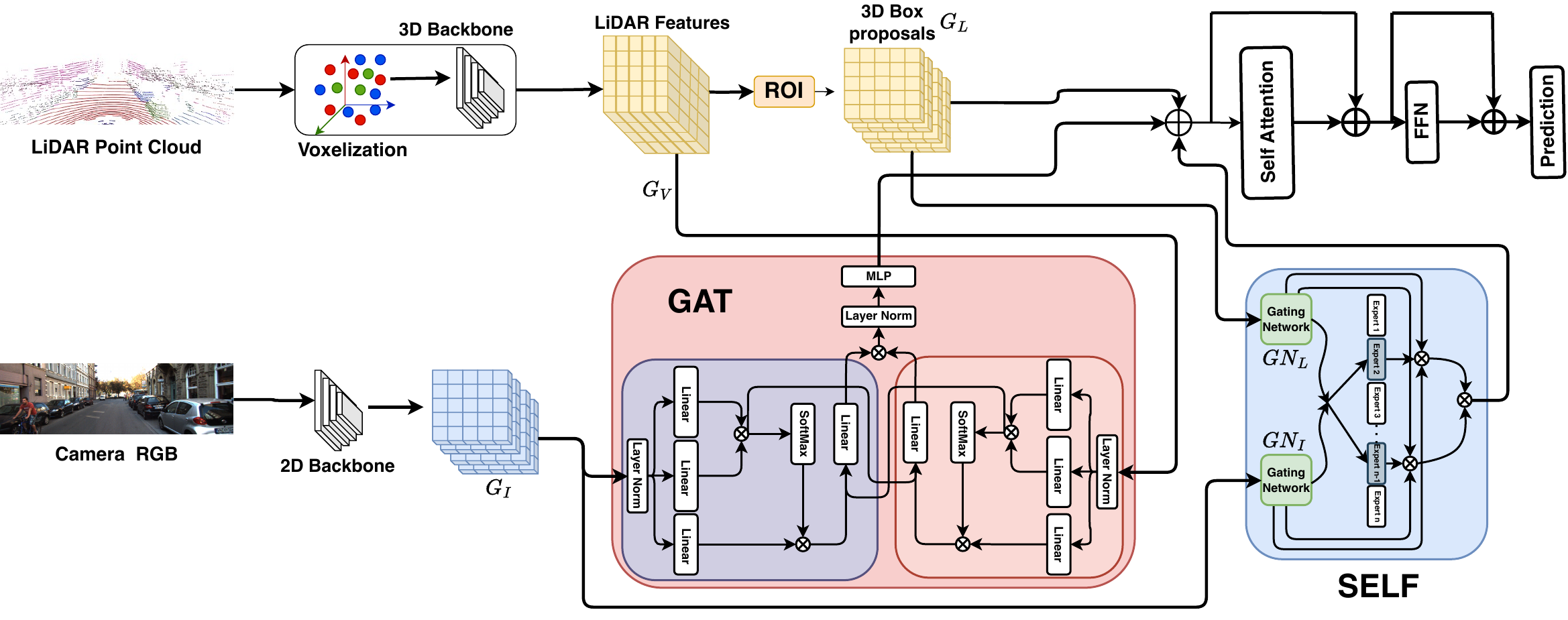}}
    \caption{Q-ICVT Pipeline: We have introduced two novel fusion blocks from extracted sparse LiDAR features ($G_V$) and dense image data ($G_I$). GAT is designed based on the adiabatic computing concept to match between the two modalities by global pointwise attention. Similarly, in SELF, the voxelized local RoI proposal feature($G_L$) is combined with a gating mechanism with $G_I$ at the local-level fusion.}
    \label{fig:cikm_main}
\end{figure*}

\textbf{Multi-modal Integration for 3D OD:}Integrating monocular vision with LiDAR point clouds enhances 3D object detection \cite{chen2016monocular, tao2023pseudo, ku2019monocular, shi2023multivariate, tao2023weakly}. Monocular systems infer 3D bounding boxes from 2D images but lack depth information \cite{tao2023pseudo}, addressed by estimating pixel-level depth \cite{tao2023pseudo}. Recognizing objects in 2D images often precedes analyzing point cloud data \cite{qi2018frustum, wang2019frustum, liu2023flatformer}, typically using a two-step, object-centered fusion approach \cite{qi2018frustum, ku2018joint}. Mid-level fusion strategies, like local-global fusion methods \cite{li2023logonet, dam2024aydiv} and other approaches \cite{piergiovanni20214d, zheng2022beyond}, combine 2D and 3D data by transferring information across their respective backbones. However, optimal alignment between camera and LiDAR features remains a major challenge \cite{li2023logonet}. Additionally, maintaining matching camera characteristics becomes complex when numerous LiDAR points are combined within a single 3D voxel \cite{dam2024aydiv}. To address this challenge, we have introduced a novel two-stage fusion method, known as Q-ICVT. Our main contributions are defined as follows:
\begin{itemize}
    \item We introduce an adiabatic computing-inspired transformer, GAT, to align sparse voxelized features with dense image features in a global context.
    \item We develop the sparse attention of gating experts, SELF, to achieve local fusion between RoI features and dense image features.
\end{itemize}

\section{Methodology}


\textbf{LiDAR and Image Feature Estimation}
Let's define the multi-modal input-output sequences as $\{(\mathcal{I}_j, {L}_j), (\mathcal{I}_{(j-1)}, {L}_{(j-1)}), \ldots\}$ for simplicity. Each input sequence at the $j$-th step consists of two types of data: LiDAR, referred to as ${L}_j$, and a camera image, represented by $x \in \mathcal{I}_j \in \mathbb{R}^{H \times W \times 3}$. The raw LiDAR point cloud for the $j$-th input is denoted as $L_{j} \rightarrow \mathcal{Q}_{j}^{raw}$, with $\mathcal{Q}_{j}^{raw} = {(\mathcal{U}_p, \mathcal{V}_p, \mathcal{W}_p, \mathcal{G}_p)}_{p=1}^{N}$, where $N$ represents the total number of points. Our objective is to design a robust local-global fusion integration to obtain adequate performance. The input point cloud data from the $j-th$ position ($\mathcal{Q}$) is converted into a voxelized representation with the coordinates $U \times V \times W \times C_V$; this representation is represented by the symbol $G_V$. The calculation of voxel features involves the mean value of point-wise features applied to non-empty voxels \cite{dam2024aydiv, li2023logonet}. The farthest point sampling (FPS) technique is employed for determining critical points \cite{shi2020pv}. This results in the generation of $\mathcal{K}$ crucial points ($G_V^{\mathcal{K}}$), where $\mathcal{K}$ is defined as 4096 for WOD. Then, the average of attributes from each point within the voxel is computed to characterize non-empty voxels, such as three-dimensional coordinates and reflectance values. Following this, a sequence of $3\times3\times3$ 3D sparse convolutions \cite{shi2019pointrcnn} are performed on the feature volumes of the point cloud. This results in the downsampling of spatial resolutions by $1 \times$, $2 \times$, $4 \times$, and $8 \times$, respectively. Following that, hierarchical intra-voxel regions (RoI) are obtained using a region proposal network \cite{dam2024aydiv, li2023logonet, shi2019pointrcnn} to generate initial bounding box proposals. Consequently, the sparse LiDAR feature is represented as \( G_V \in \mathbb{R}^{{H_V} \times {W_V} \times C_V} \). Similarly, dense semantic image features \( G_I \in \mathbb{R}^{{H_I} \times {W_I} \times C_I} \) are obtained using a 2D detector \cite{ren2015faster, liu2021swin}.

\subsection{Adiabatic computing in GAT}

In the context of  vision transformers \cite{zhang2021multi,hatamizadeh2023global}, the main goal is to utilize the reversibility of adiabatic processes \cite{ciliberto2018quantum, rajesh2021quantum, rauchenecker2017exploiting, li1996reversibility} to improve the global context and effectively handle the interaction between two modalities: LiDAR point cloud ($G_V \in \mathbb{R}^{{H_V} \times {W_V} \times C_V}$) and image data ($G_I \in \mathbb{R}^{{H_I} \times {W_I} \times C_I}$). In order to assess the reversibility process, both forward and backward transformations were conducted between two distinct feature modalities. One modality was represented in a sparse form using LiDAR, while the other modality was represented in a dense form retrieved from the input image .




Let us consider a reversible block in a transformer layer, as illustrated in Figure \ref{fig:cikm_main}. The GAT module is capable of effectively handling sparse and dense modalities through forward and reverse mechanisms \cite{harwood2022improving}. To attain this objective, we first consider the forward transition through linear projection data as a global query \cite{hatamizadeh2023global} with key-value pair matching. These forward mechanism transformer blocks help retain not only the focus on the central forward points around the voxelized sparse feature ($G_V$), but also identify affine matrices within the dense image ($G_I$). Unlike existing methods \cite{li2023logonet, ma2023detzero}, the global spatial resolution around the voxelized LiDAR point feature ($G_V$) does not guarantee a comprehensive sense of the dense image feature ($G_I$). Therefore, we also introduce another reversible transformer block for query-key-value matching of the voxelized LiDAR feature \cite{mangalam2022reversible}, which will be considered as the backward transition of the matrix. Therefore, the forward transformation block in GAT is defined as follows:

\begin{align}\label{eq:forward}
  \mathbf{Y}_f = \mathcal{F}(G_I)  
\end{align}
where $\mathcal{F} \in \mathbb{R}^{{H_I} \times {W_I} \times C_I}$ is the forward transformer block \cite{hatamizadeh2022global}. Here,$ \mathbf{Q}, \mathbf{K}, \mathbf{V} \in G_I$ and we use layer normalization in between the query-key-value matching \cite{hatamizadeh2023global}. Similarly, the backward transformation for the voxelized LiDAR feature is represented as follows:
\begin{align}\label{eq:forward}
  \mathbf{Y}_r = \mathcal{F}^{-1}(G_V)  
\end{align}
where \(\mathcal{F}^{-1} \in \mathbb{R}^{{H_V} \times {W_V} \times C_V}\) is the backward transformer block \cite{hatamizadeh2022global}. Here, \(\mathbf{Q}, \mathbf{K}, \mathbf{V} \in G_V\), and we use layer normalization similar to the forward path \cite{mangalam2022reversible}.
To accumulate the dimension matching, we concatenate the matrix transition through a linear layer. Therefore, we concatenate the forward and reverse blocks in GAT using the $\bigoplus$  operation. Finally, the GAT model is obtained as follows:
\begin{align}\label{eq:GAT}
G_{VI} = \mathcal{G}(\mathbf{Y}_f \bigoplus \mathbf{Y}_r )
\end{align}
where \(\bigoplus\) denotes the concatenation operation. \(G_{VI} \in \mathbb{R}^{{H_I} \times {W_I} \times C_I}\) ensures that the original dense image feature (\(G_I\)) can be retrieved.

\subsection{Sparse Expert of Local Fusion (SELF)}
To achieve local-level fusion, we introduce Sparse Expert fusion for voxelized LiDAR data. A region proposal network \cite{dam2024aydiv, li2023logonet, shi2019pointrcnn} generates initial bounding box suggestions (\( P = \{P_1, P_2, \ldots, P_n\} \)) based on multi-level voxel features (\(G_V\)). The multi-level voxel features of RoI are defined by \(G_L\). We adopt the Mixture of Experts (MoE) \cite{shazeer2017outrageously, riquelme2021scaling} model, known for capturing long dependencies in heterogeneous sequence datasets \cite{han2024fusemoe, yang2024worldgpt, dam2021improving}, for sparse RoI LiDAR features (\(G_L\)) and dense image features (\(G_I\)). This robust framework leverages specialized sub-models or "experts," each optimized for different input subsets. We extend two separate gating networks (A gating network determines the weights for each expert), \( GN_{\text{L}}(.) \) for LiDAR and \( GN_{\text{I}}(.) \) for image data, to determine the weights for each gating networks. For a given LiDAR input \( G_{\text{L}} \) and an image input \( G_{\text{I}} \), the outputs of both modalities are computed as:
\begin{align}\label{eq:_gate_yl}
    y_{L} = \sum_{i=1}^{N} {GN_{Li}}(G_{L}) {E_{Li}} (G_{L})
\end{align}

\begin{align}\label{eq:_gate_yI}
    y_{I} = \sum_{i=1}^{N} {GN_{Ii}}(G_{I}) {E_{Ii}} (G_{I})
\end{align}
where, $y_{L}$ and $y_{I}$ are each gating mechanism outputs for each modalities passing through by selecting by expert networks. $N$ numbers of experts for each modalities. Therefore, the gating network($GN$) is defined as follows, 

\begin{align}\label{eq:gating}
GN(x) = \operatorname{Softmax}(\operatorname{TopK}(\mathbb{H}_w(x), k))
\end{align}
where, $\mathbb{H}_w(x)$ is the individual gating function that can be represented by the parametric weight $w$, $\mathbb{H}_w(x) = \Psi(  \operatorname{Softplus}\left(x \cdot \delta\right))$, where $\delta$ is the noise \cite{lin2024moe}.

$E(.)$ is an expert network function that will choose top $K$ values. Therefore, selecting $\text{Top}K$ experts from  $E(.)$ is represented as follows \cite{dam2022developing},


\begin{align}\label{eq:expert}
\{E_{L/I, k}\}_{k=1}^N =
\begin{cases}
E_{L/I, j} & \text{if } j \in \argmaxA_{1:K} \{GN_{L/I, j}\}_{j=1}^N \\
-\infty & \text{otherwise}
\end{cases}
\end{align}

These dual-gating mechanism (Equ. \ref{eq:expert} and Equ. \ref{eq:gating})allows the model to independently assess and integrate the specific characteristics of each modality. By doing so, the model can more effectively capture the complementary information provided by LiDAR and image data. The final fused output \( y \) is then derived by combining \( y_{\text{L}} \) and \( y_{\text{I}} \) through a subsequent fusion network \( \mathbb{F} \):
\begin{align}\label{eq:self}
    y = \mathbb{F}(y_L, y_I)
\end{align}
where \(\mathbb{F}\) can be easily obtained through an embedding function. This method fuses distinct data types to boost multimodal task performance.

\section{Experimental Validation}
\subsection{Dataset Details}
\textbf{WOD} achieves excellent performance in 3D object detection benchmarks, thanks to its extensive dataset consisting of more than 200,000 frames, 1,150 sequences, and a combination of LiDAR, images \cite{sun2020scalability}. The dataset comprises 798 training sequences, 202 validation sequences, and 150 testing sequences. The detection range is 75 meters, and the coverage area is 150 meters by 150 meters. We evaluate models using Average Precision (AP) and Average Precision weighted by Heading (APH) as described in  \cite{sun2020scalability, li2023logonet}. We include results for both \text{LEVEL\texttt{\string_}1 (L1)} and \text{LEVEL\texttt{\string_}2 (L2)} difficulty items, offering a thorough assessment and comparison of the models' performance.
\subsection{Evolution of WOD performance}
\begin{table*}
\centering
\caption{Evaluation of Model Performance for 3D Detection on the WOD Test Set. In this table, `L' represents LiDAR sensors, `I' represents camera sensors, `TTA' stands for test-time augmentation, and `Ens' denotes ensemble model outputs, marked by \texttt{\#}.}
\label{tab:AYDIV_test_set}
\scalebox{0.85}{
\begin{tabular}{l|l|l|ll|ll|ll}
\hline
\textbf{Method} & \textbf{Modality} & \textbf{ALL (mAPH)} & \multicolumn{2}{l|}{\textbf{VEH (AP/APH)}} & \multicolumn{2}{l|}{\textbf{PED (AP/APH)}} & \multicolumn{2}{l|}{\textbf{CYC (AP/APH)}} \\ \hline
 &  & L2 & \multicolumn{1}{l|}{L1} & L2 & \multicolumn{1}{l|}{L1} & L2 & \multicolumn{1}{l}{L1} & L2 \\ \hline

 Q-ICVT Ens (ours) \texttt{\#}& L+I & \textcolor{blue}{82.54} \textcolor{red}{(+1.52)}& \multicolumn{1}{l|}{\textcolor{blue}{89.21}/\textcolor{blue}{88.98}} & \textcolor{blue}{83.46}/\textcolor{blue}{82.69} & \multicolumn{1}{l|}{\textcolor{blue}{89.01}/\textcolor{blue}{86.43}} & \textcolor{blue}{85.92}/\textcolor{blue}{83.82} & \multicolumn{1}{l|}{\textcolor{blue}{84.71}/\textcolor{blue}{84.01}} & \textcolor{blue}{82.41}/\textcolor{blue}{81.12} \\ \hline
LoGoNet Ens\texttt{\#} \cite{li2023logonet}  & L+I & 81.02 & \multicolumn{1}{l|}{88.33/87.87} & 82.17/81.72 & \multicolumn{1}{l|}{88.98/85.96} & 84.27/81.28 & \multicolumn{1}{l|}{83.10/82.16} & 80.93/80.06 \\ 
BEVFusion TTA\texttt{\#} \cite{liu2023bevfusion} & L+I & 79.97 & \multicolumn{1}{l|}{87.96/87.58} & 81.29/80.92 & \multicolumn{1}{l|}{87.64/85.04} & 82.19/79.65 & \multicolumn{1}{l|}{82.53/81.67} & 80.17/79.33 \\ 
LidarMultiNet TTA\texttt{\#} \cite{ye2023lidarmultinet} & L & 79.94 & \multicolumn{1}{l|}{87.64/87.26} & 80.73/80.36 & \multicolumn{1}{l|}{87.75/85.07} & 82.48/79.86 & \multicolumn{1}{l|}{82.77/81.84} & 80.50/79.59 \\ 
MPPNet Ens\texttt{\#} \cite{chen2022mppnet} & L & 79.60 & \multicolumn{1}{l|}{87.77/87.37} & 81.33/80.93 & \multicolumn{1}{l|}{87.92/85.15} & 82.86/80.14 & \multicolumn{1}{l|}{80.74/79.90} & 78.54/77.73 \\ 
MT-Net Ens\texttt{\#} \cite{chen2022mt} & L & 78.45 & \multicolumn{1}{l|}{87.11/86.69} & 80.52/80.11 & \multicolumn{1}{l|}{86.50/83.55} & 80.95/78.08 & \multicolumn{1}{l|}{80.50/79.43} & 78.22/77.17 \\ 
DeepFusion Ens\texttt{\#} \cite{li2022deepfusion} & L+I & 78.41 & \multicolumn{1}{l|}{86.45/86.09} & 79.43/79.09 & \multicolumn{1}{l|}{86.14/83.77} & 80.88/78.57 & \multicolumn{1}{l|}{80.53/79.80} & 78.29/77.58 \\ 
AFDetV2 Ens\texttt{\#} \cite{hu2022afdetv2} & L & 77.64 & \multicolumn{1}{l|}{85.80/85.41} & 78.71/78.34 & \multicolumn{1}{l|}{85.22/82.16} & 79.71/76.75 & \multicolumn{1}{l|}{81.20/80.30} & 78.70/77.83 \\ 
INT Ens\texttt{\#} \cite{xu2022int} & L & 77.21 & \multicolumn{1}{l|}{85.63/85.23} & 79.12/78.73 & \multicolumn{1}{l|}{84.97/81.87} & 79.35/76.36 & \multicolumn{1}{l|}{79.76/78.65} & 77.62/76.54 \\ 
HoriLiDAR3D Ens\texttt{\#} \cite{ding20201st} & L+I & 77.11 & \multicolumn{1}{l|}{85.09/84.68} & 78.23/77.83 & \multicolumn{1}{l|}{85.03/82.10} & 79.32/76.50 & \multicolumn{1}{l|}{79.73/78.78} & 77.91/76.98 \\ \hline
Q-ICVT (ours)  & L+I & \textcolor{blue}{78.34} \textcolor{red}{(+1.24)} & \multicolumn{1}{l|}{\textcolor{blue}{87.73}/\textcolor{blue}{87.24}} & {\textcolor{blue}{80.84}/\textcolor{blue}{80.10}} & \multicolumn{1}{l|}{\textcolor{blue}{87.96}/\textcolor{blue}{85.53}} & {\textcolor{blue}{82.89}/\textcolor{blue}{80.47}} & \multicolumn{1}{l|}{\textcolor{blue}{77.86}/\textcolor{blue}{76.33}} & {\textcolor{blue}{74.87}/\textcolor{blue}{74.45}} \\  \hline
LoGoNet \cite{li2023logonet}  & L+I & 77.10 & \multicolumn{1}{l|}{86.51/86.10} & 79.69/79.30 & \multicolumn{1}{l|}{86.84/84.15} & 81.55/78.91 & \multicolumn{1}{l|}{76.06/75.25} & 73.89/73.10 \\ 
BEVFusion \cite{liu2023bevfusion} & L+I & 76.33 & \multicolumn{1}{l|}{84.97/84.55} & 77.88/77.48 & \multicolumn{1}{l|}{84.72/81.97} & 79.06/76.41 & \multicolumn{1}{l|}{78.49/77.54} & 76.00/75.09 \\ 
CenterFormer \cite{zhou2022centerformer} & L & 76.29 & \multicolumn{1}{l|}{85.36/84.94} & 78.68/78.28 & \multicolumn{1}{l|}{85.22/82.48} & 80.09/77.42 & \multicolumn{1}{l|}{76.21/75.32} & 74.04/73.17 \\ 
MPPNet \cite{chen2022mppnet} & L & 75.67 & \multicolumn{1}{l|}{84.27/83.88} & 77.29/76.91 & \multicolumn{1}{l|}{84.12/81.52} & 78.44/75.93 & \multicolumn{1}{l|}{77.11/76.36} & 74.91/74.18 \\ 
DeepFusion \cite{li2022deepfusion} & L+I & 75.54 & \multicolumn{1}{l|}{83.25/82.82} & 76.11/75.69 & \multicolumn{1}{l|}{84.63/81.80} & 79.16/76.40 & \multicolumn{1}{l|}{77.81/76.82} & 75.47/74.51 \\ \hline
\end{tabular}
}
\label{tab:Waymo_test_set}
\end{table*}

\begin{table*}
\centering
\caption{Comparative Performance Analysis on the Waymo Validation Set for 3D Vehicle Detection (IoU = 0.7), Pedestrian Detection (IoU = 0.5), and Cyclist Detection (IoU = 0.5). PV-RCNN \cite{shi2020pv} is our baseline model. }
\scalebox{0.85}{
\begin{tabular}{l|l|l|ll|ll|ll}
\hline
\textbf{Method} & \textbf{Modality} & \textbf{ALL (mAPH)} & \multicolumn{2}{l|}{\textbf{VEH (AP/APH)}} & \multicolumn{2}{l|}{\textbf{PED (AP/APH)}} & \multicolumn{2}{l|}{\textbf{CYC (AP/APH)}} \\ \hline
 &  & L2 & \multicolumn{1}{l|}{L1} & L2 & \multicolumn{1}{l|}{L1} & L2 & \multicolumn{1}{l|}{L1} & L2 \\ \hline
SECOND \cite{yan2018second} & L & 57.23 & 72.27/71.69 & 63.85/63.33 & 68.70/58.18 & 60.72/51.31 & 60.62/59.28 & 58.34/57.05 \\ 
PointPillars \cite{lang2019pointpillars} & L & 57.53 & 71.60/71.00 & 63.10/62.50 & 70.60/56.70 & 62.90/50.20 & 64.40/62.30 & 61.90/59.90 \\ 
LiDAR-RCNN \cite{li2021lidar} & L & 60.10 & 73.50/73.00 & 64.70/64.20 & 71.20/58.70 & 63.10/51.70 & 68.60/66.90 & 66.10/64.40 \\ 

CenterPoint\cite{yin2021center} & L & 65.46 & - & -/66.20 & - & -/62.60 & - & -/67.60 \\ 
PointAugmenting \cite{wang2021pointaugmenting} & L+I & 66.70 & 67.4/- & 62.7/- & 75.04/- & 70.6/- & 76.29/- & 74.41/- \\ 
Pyramid-PV \cite{mao2021pyramid} & L & - & 76.30/75.68 & 67.23/66.68 & - & - & - & - \\ 
PDV \cite{hu2022point} & L & 64.25 & 76.85/76.33 & 69.30/68.81 & 74.19/65.96 & 65.85/58.28 & 68.71/67.55 & 66.49/65.36 \\ 
Graph-RCNN \cite{yang2022graph} & L & 70.91 & 80.77/80.28 & 72.55/72.10 & 82.35/76.64 & 74.44/69.02 & 75.28/74.21 & 72.52/71.49 \\ 
3D-MAN \cite{yang20213d} & L & - & 74.50/74.00 & 67.60/67.10 & 71.70/67.70 & 62.60/59.00 & - & - \\ 
Centerformer \cite{zhou2022centerformer} & L & 73.70 & 78.80/78.30 & 74.30/73.80 & 82.10/79.30 & 77.80/75.00 & 75.20/74.40 & 73.20/72.30 \\ 
DeepFusion \cite{li2022deepfusion} & L+I & - & 80.60/80.10 & 72.90/72.40 & 85.80/83.00 & 78.70/76.00 & - & - \\ 
MPPNet \cite{chen2022mppnet} & L & 74.22 & 81.54/81.06 & 74.07/73.61 & 84.56/81.94 & 77.20/74.67 & 77.15/76.50 & 75.01/74.38 \\ 
MPPNet \cite{chen2022mppnet} & L & 74.85 & 82.74/82.28 & 75.41/74.96 & 84.69/82.25 & 77.43/75.06 & 77.28/76.66 & 75.13/74.52 \\ 
LoGoNet \cite{li2023logonet} & L+I & 75.54 & 83.21/82.72 & 75.84/75.38 & 85.80/83.14 & 78.97/76.33 & 78.58/77.79 & 75.67/74.91 \\ \hline
Baseline\cite{shi2020pv} & L & 63.33 & 77.51/76.89 & 68.98/68.41 & 75.01/65.65 & 66.04/57.61 & 67.81/66.35 & 65.39/63.98 \\ 
Q-ICVT (ours)  & L+I & \textcolor{blue}{78.21} \textcolor{red}{(+14.88)} & \textcolor{blue}{86.12}/\textcolor{blue}{84.96} & \textcolor{blue}{81.32}/\textcolor{blue}{79.37} & \textcolor{blue}{88.50}/\textcolor{blue}{85.94} & \textcolor{blue}{82.68}/\textcolor{blue}{78.69} & \textcolor{blue}{81.54}/\textcolor{blue}{80.56} & \textcolor{blue}{77.94}/\textcolor{blue}{76.57} \\ \hline
\end{tabular}
}
\label{tab:Waymo_valid_set}
\end{table*}

The detailed performance of Q-ICVT, both single and ensemble variants, on the WOD test and validation sets is presented in Table \ref{tab:Waymo_test_set} and Table \ref{tab:Waymo_valid_set}. According to Table \ref{tab:Waymo_test_set}, Q-ICVT excels, surpassing other leading methods in both L1 and L2 difficulty levels. Compared to LoGoNet \cite{li2023logonet}, Q-ICVT shows significant gains, especially in L2 difficulty with a 1.24 increase in APH value, achieved without ensemble techniques. Specifically, the non-ensemble version of Q-ICVT outperforms LoGoNet \cite{li2023logonet} by 1.22 AP/L1, 1.14 APH/L1, 1.15 AP/L2, and 0.80 APH/L2 for vehicles; 1.12 AP/L1, 1.38 APH/L1, 1.34 AP/L2, and 1.56 APH/L2 for pedestrians; and 1.80 AP/L1, 1.08 APH/L1, 0.98 AP/L2, and 1.35 APH/L2 for cyclists, leading to a total improvement of 1.24 mAPH/L2.The ensemble version of Q-ICVT also surpasses LoGoNet-Ens \cite{li2023logonet} with differences of 0.88 AP/L1, 1.11 APH/L1, 1.29 AP/L2, and 0.97 APH/L2 for vehicles; 0.03 AP/L1, 0.47 APH/L1, 1.65 AP/L2, and 2.54 APH/L2 for pedestrians; and 1.61 AP/L1, 1.85 APH/L1, 1.48 AP/L2, and 1.06 APH/L2 for cyclists, resulting in an overall improvement of 1.52 mAPH/L2.


Table \ref{tab:Waymo_valid_set} presents an extensive comparison of model performance for 3D object detection on the WOD validation set. Remarkably, Q-ICVT shows substantial advancements across various difficulty levels. For L1 difficulty, it surpasses the validation results of LoGoNet \cite{li2023logonet} on WOD by 2.91 AP/L1, 2.24 APH/L1, 5.48 AP/L2, and 3.99 APH/L2 for vehicles; 2.70 AP/L1, 2.80 APH/L1, 3.71 AP/L2, and 2.36 APH/L2 for pedestrians; and 2.96 AP/L1, 2.77 APH/L1, 2.27 AP/L2, and 1.66 APH/L2 for cyclists, resulting in an overall improvement of 2.67 mAPH/L2. These improvements highlight Q-ICVT proficiency in accurately detecting all classes, emphasizing the effectiveness of multi-modal feature alignment in enhancing 3D object detection.

\section{Ablation Studies on WOD}

\textbf{ Influence of each component.} Table \ref{tab:each_component} highlights the effects of individual components on L2 difficulty in the QICVT WOD test set. When only the Global Adiabatic Transformer (GAT) is used without the Sparse Expert of Local Fusion (SELF), performance decreases to 79.56 for vehicles, 80.12 for pedestrians, and 78.97 for cyclists. This drop is due to the limitations of the SELF component, which fails to achieve optimal fusion despite integrating voxel RoI features into the GAT module. The exclusion of GAT leads to an even greater decline in performance: 78.27 for vehicles, 79.51 for pedestrians, and 77.34 for cyclists. Therefore, even though local fusion centroids are closer to the image surface, voxel point centroids are not able to provide dense image feature information, diminishing the effectiveness of global cross-modal fusion. Combining both GAT and SELF components results in significant performance improvements, with scores of 82.69 for vehicles, 83.82 for pedestrians, and 81.12 for cyclists, underscoring the importance of both components for optimal performance.
\begin{table}[h]
\centering
\caption{Each component's effect on L2 difficulty in the Q-ICVT WOD test set}
\begin{tabular}{ll|lll} 
\hline
\multicolumn{2}{l}{Components} & \multicolumn{3}{l}{APH (L2)} \\ 
\hline
GAT & SELF & VEH & PED & CYC \\  \hline

\checkmark & & 79.56 & 80.12 & 78.97 \\ 
& \checkmark & 78.27 & 79.51 & 77.34 \\ 
\checkmark & \checkmark & \textcolor{blue}{82.69} & \textcolor{blue}{83.82} & \textcolor{blue}{81.12} \\ 
\hline
\end{tabular}
\label{tab:each_component}
\end{table}

\section{Conclusion}
We introduced QICVT, a 3D multi-modal object detection method based on transformers, consisting of two key components:  GAT and SELF. These components were designed to capture both local and global dependencies, thereby enhancing the efficacy of 3D detection at both short and long distances.  To determine the efficacy of QICVT, we conducted comprehensive experiments on the WOD  benchmark datasets. QICVT demonstrated its efficacy in multi-modal object detection by achieving competitive performance when compared to state-of-the-art methods. In addition, we conducted comprehensive ablation experiments to compare the effect of each proposed component on QICVT’s performance.

\bibliographystyle{ACM-Reference-Format}
\bibliography{sample-base}

\end{document}